\documentclass[conference]{IEEEtran}

\usepackage[utf8]{inputenc}
\usepackage[english]{babel}
\usepackage[T1]{fontenc}
\usepackage{cite}
\usepackage{amsmath,amssymb,amsfonts}
\usepackage{algorithmic}
\usepackage{graphicx}
\usepackage{textcomp}
\usepackage{xcolor}
\usepackage{multirow}
\usepackage{array} 
\usepackage{makecell} 
\usepackage{float}

\def\BibTeX{{\rm B\kern-.05em{\sc i\kern-.025em b}\kern-.08em T\kern-.1667em\lower.7ex\hbox{E}\kern-.125em}}

\begin{document}
	
	\title{Comparison of Image Preprocessing Techniques for Vehicle License Plate Recognition Using OCR: Performance and Accuracy Evaluation}
	
	\author{\IEEEauthorblockN{Renato Augusto Tavares}
		\IEEEauthorblockA{\textit{Instituto de Informática} \\
			\textit{Universidade Federal de Goiás} \\
			Goiânia, Goiás \\
			rat@discente.ufg.br}}
	
	\maketitle
	
	\begin{abstract}
		The growing use of Artificial Intelligence solutions has led to an explosion in image capture and its application in machine learning models. However, the lack of standardization in image quality generates inconsistencies in the results of these models. To mitigate this problem, Optical Character Recognition (OCR) is often used as a preprocessing technique, but it still faces challenges in scenarios with inadequate lighting, low resolution, and perspective distortions.
		
		This work aims to explore and evaluate various preprocessing techniques, such as grayscale conversion, CLAHE in RGB, and Bilateral Filter, applied to vehicle license plate recognition. Each technique is analyzed individually and in combination, using metrics such as accuracy, precision, recall, F1-score, ROC curve, AUC, and ANOVA, to identify the most effective method. The study uses a dataset of Brazilian vehicle license plates, widely used in OCR applications. The research provides a detailed analysis of best preprocessing practices, offering insights to optimize OCR performance in real-world scenarios.
	\end{abstract}
	
	\begin{IEEEkeywords} 
		ocr, optical character recognition, vehicle license plates, image preprocessing, artificial intelligence
	\end{IEEEkeywords}
	
	\section{Introduction}
	With the increase in the use of Artificial Intelligence solutions, the capture of images and their subsequent use in machine learning models have grown exponentially. However, the quality and technique used to capture these images do not follow a standard, resulting in Artificial Intelligence models generating inconsistent or incoherent results. To mitigate this problem, researchers commonly resort to OCR (Optical Character Recognition) techniques to preprocess the images before using them in machine learning models, ensuring greater accuracy in the results.
	
	The initial concept of OCR was developed in the 1920s by Emanuel Goldberg, but it wasn't until the 1970s that more advanced techniques began to emerge, with contributions from Ray Kurzweil. Despite over 50 years of technological evolution, OCR still faces significant challenges in pattern detection, especially when certain conditions are not ideal. Factors such as inadequate lighting, low resolution, noise in images, and distortions caused by perspective or angle continue to hinder OCR performance. These limitations are particularly evident in real-world scenarios such as vehicle license plate recognition, where systems must deal with a wide variety of environmental conditions.
	
	The objective of this work is, aware of the mentioned limitations, to address research questions related to the current state of the art in OCR and to test various image preprocessing techniques, such as Grayscale Conversion, CLAHE (Contrast Limited Adaptive Histogram Equalization) in RGB images, Bilateral Filter, and others. These techniques will be evaluated individually and in combination to determine which ones offer the best results in terms of accuracy and efficiency in license plate recognition. To ensure the validity of the results, we use statistical metrics such as accuracy, precision, recall, F1-score, ROC curve, AUC, and ANOVA, allowing for a rigorous quantitative analysis of the performance of each preprocessing technique.
	
	To make the work replicable, we opted to use a dataset of Brazilian vehicle license plates, a common application for OCR. License plate detection is widely explored in studies involving neural networks, as shown in recent works such as "Comparative Analysis of EasyOCR and TesseractOCR for Automatic License Plate Recognition" \cite{b1} and "Real-Time License Plate Detection and Recognition System using YOLOv7x and EasyOCR" \cite{b2}. These papers demonstrate how different OCR techniques can be applied to license plate recognition in real conditions and serve as a basis for our choice of dataset and methodology.
	
	Furthermore, the choice to work with DSLR camera captures at different distances and lighting conditions allows us to test how OCR performs in scenarios where image quality can vary significantly, as discussed in "Deep Learning Model for Automatic Number License Plate Detection and Recognition System" \cite{b3}. The use of different focal distances is also aligned with research on capturing plates at different angles and distances, enabling a comprehensive analysis of the impact of these variations on OCR performance.
	
	Thus, this study aims not only to identify the most effective preprocessing technique but also to provide a set of best practices for researchers working with license plate recognition and other OCR applications. By combining a rigorous validation process with a realistic dataset and advanced preprocessing techniques, we hope to offer a significant contribution to the field of Optical Character Recognition, with potential impact on traffic monitoring, vehicle security, and urban automation industries.
	
	\section{Research Methodology and Research Questions}
	Systematic literature reviews have gained increasing relevance in academic research, as they follow a structured, rigorous, and reliable methodology for searching and analyzing publications. This type of review allows readers to quickly and effectively obtain a panoramic view of the field of study \cite{b4}. Used to deepen knowledge about specific areas, systematic reviews also allow for the identification of gaps and opportunities for future research. The main objective of this approach is to locate, interpret, evaluate, and classify all relevant articles to the predefined research questions.
	
\begin{figure*}[htbp]
		\centerline{\includegraphics[width=\textwidth]{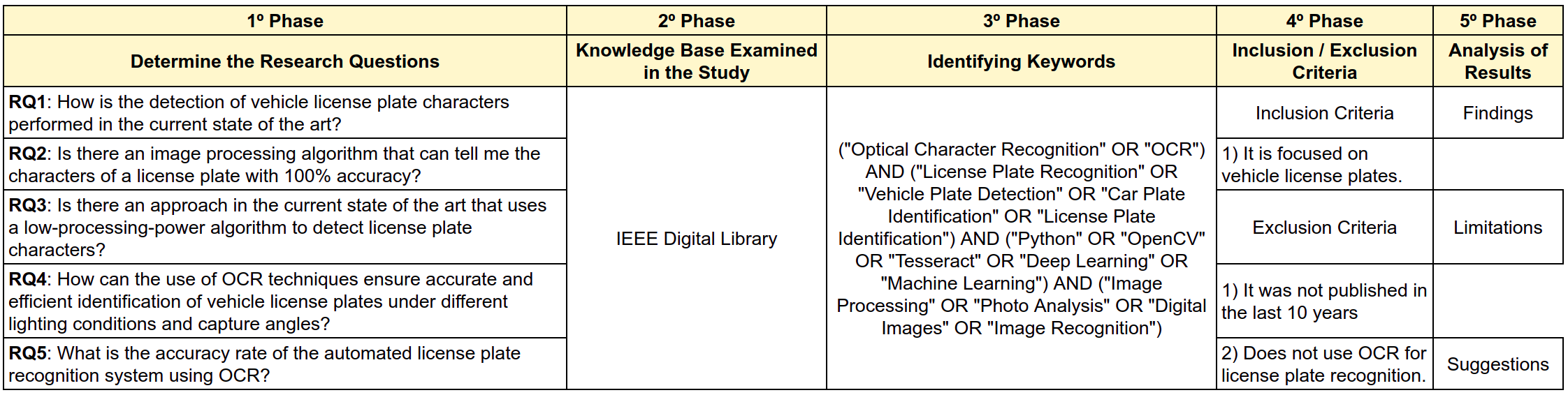}}
		\caption{Systematic literature review stages.}
		\label{img1}
	\end{figure*}

	In this study, we adopted a systematic review process in five stages, aiming to map the existing literature, define research questions, and identify new relevant keywords. Figure~\ref{img1} presents the phases followed during the review, while Table~\ref{tab1} illustrates the number of articles analyzed throughout the process.
	
	During the initial selection, using strategically defined keywords, the \textbf{IEEE Digital Library} returned a total of \textbf{190 articles}. As part of the rigorous filtering process, we applied our exclusion criteria to ensure the relevance and suitability of the studies, resulting in the elimination of 42 articles, leaving a total of 148 for analysis. In the subsequent stage, we removed duplicate articles, identified either by human error or by internal failures of the search tool, resulting in the exclusion of two more articles, leaving \textbf{146 articles} that were carefully read and analyzed for the construction of this work.
	
	\begin{table}[H]
		\caption{Stages of Systematic Literature Review}
		\begin{center}
			\begin{tabular}{l|c|c|c|c|}
				\cline{2-5}
				& Start & Step 1 & Step 2 & Step 3 \\ \hline
				\multicolumn{1}{|l|}{Initial Selection}       & 190    &         &         &         \\ \hline
				\multicolumn{1}{|l|}{Exclusion Criteria} &        & -42     &         &         \\ \hline
				\multicolumn{1}{|l|}{Duplicate Removal} &        &         & -2      &         \\ \hline
				\multicolumn{1}{|l|}{Final Result}       &        &         &         & 146     \\ \hline
			\end{tabular}
			\label{tab1}
		\end{center}
	\end{table}
	
	This systematic survey was conducted with great care, aiming to ensure that we were not only informed about the state of the art in the area of optical character recognition (OCR) and image processing but also that we could accurately identify the main open research questions. Each step of the process was designed to provide a comprehensive view of the current trends and challenges in the field, ensuring that our research was aligned with the latest scientific developments.
	
	Through this approach, we were able not only to map the most relevant contributions but also to identify gaps in the literature, allowing us to formulate a work that genuinely contributes to advancing the field. The care in ensuring that the articles were appropriate for our study theme was essential to guarantee that we addressed the main research questions in an innovative and rigorous way, consolidating our work on a solid scientific basis.
	
	The research questions were developed to define, precisely, the purpose of the study, considering its scope. The research questions guiding this work are as follows:
	
	\begin{itemize}
		\item \textbf{RQ1:} How is the detection of vehicle license plate characters done in the current state of the art?
		\item \textbf{RQ2:} Is there an image processing algorithm that provides 100
		\item \textbf{RQ3:} How can the use of OCR techniques ensure accurate and efficient identification of vehicle license plates under different lighting conditions and capture angles?
		\item \textbf{RQ4:} What is the accuracy rate of an automated license plate recognition system using OCR?
	\end{itemize}
	
	\section{Statistical Evaluation Metrics Used}
	
	As mentioned in the introduction, it is essential to thoroughly understand the image preprocessing algorithms to significantly improve the accuracy of character recognition systems. The effective training of Artificial Intelligences (AI), as well as the development of advanced technologies like autonomous vehicles, automatic traffic license plate identification systems, and drone-based automated flights, depend on highly accurate and reliable OCR (Optical Character Recognition) algorithms. Any error in these systems can not only cause financial loss but also put lives at risk, underscoring the importance of improving these processes.
	
	In this section, we will deeply discuss the five preprocessing algorithms applied in this work, detailing their advantages, disadvantages, and the scenarios in which each one performs best, as well as those where its use is less efficient. The goal is to establish a robust statistical metrics base, providing a clear and grounded quantitative analysis that can serve as a reference for researchers and developers looking to train AIs using OCR.
	
	Among the algorithms used, techniques such as grayscale conversion and the use of CLAHE (Contrast Limited Adaptive Histogram Equalization) in RGB images stand out, widely recognized for improving contrast and the visibility of important details, as demonstrated in previous studies on vehicle license plate recognition [1]. Additionally, the Bilateral Filter was tested, known for its effectiveness in noise reduction without compromising image edges, which is particularly relevant in unfavorable lighting situations or damaged plates.
	
	Previous studies, such as "Comparative Analysis of EasyOCR and TesseractOCR for Automatic License Plate Recognition" \cite{b1} and "Deep Learning Model for Automatic Number License Plate Detection and Recognition System" \cite{b2}, have already demonstrated the importance of these techniques in improving accuracy in OCR systems, especially when applied to plates captured under varied conditions, such as different angles, variable distances, and uneven lighting. By exploring these approaches, this study seeks to determine which preprocessing technique offers the best results in terms of accuracy, precision, recall, and F1-score, as well as identifying the limitations of each method in specific situations.
	
	At the end of this analysis, we aim to offer a viable and scientifically grounded solution for researchers and engineers who need to optimize character recognition in their AI applications, whether they are focused on traffic monitoring systems, vehicular automation, or other critical scenarios where accuracy is essential for safety and efficiency.
	
	To facilitate the understanding of the article, let's define some metrics used in the text.
	
	\subsection{Accuracy}
	
	\textbf{Accuracy} is an evaluation metric used to measure the effectiveness of a classification model. It represents the fraction of correct predictions out of the total number of predictions made by the model. Accuracy is an important metric in cases where the classes are balanced, meaning the number of positive and negative cases is approximately the same.
	
	The formula to calculate accuracy is:
	
	\[
	\text{Accuracy} = \frac{TP + TN}{TP + TN + FP + FN}
	\]
	
	Where:
	
	\begin{itemize}
		\item \textbf{TP} (True Positives) are correct predictions for the positive class (correctly predicted plates).
		\item \textbf{TN} (True Negatives) are correct predictions for the negative class (plates that were not incorrectly predicted).
		\item \textbf{FP} (False Positives) are incorrect predictions for the positive class (incorrect plates predicted as correct).
		\item \textbf{FN} (False Negatives) are incorrect predictions for the negative class (correct plates predicted as incorrect).
	\end{itemize}
	
	\subsection{Precision}
	
	\textbf{Precision} is an evaluation metric that indicates the proportion of correct positive predictions out of the total number of positive predictions made by the model. Precision is useful in situations where the cost of false positives is high, meaning it is important to minimize positive prediction errors.
	
	The formula for precision is:
	
	\[
	\text{Precision} = \frac{TP}{TP + FP}
	\]
	
	Where:
	
	\begin{itemize}
		\item \textbf{TP} (True Positives) are correct predictions for the positive class (correctly predicted plates).
		\item \textbf{FP} (False Positives) are incorrect predictions for the positive class (plates predicted as correct, but that are incorrect).
	\end{itemize}
	
	\subsection{Recall (Sensitivity)}
	
	\textbf{Recall}, also known as sensitivity or true positive rate, measures the model's ability to correctly identify positive instances (in this case, correctly predicted plates). Recall is particularly important when the goal is to minimize false negatives, meaning when it is crucial to detect all positive cases.
	
	The formula for \textbf{Recall} is:
	
	\[
	\text{Recall} = \frac{TP}{TP + FN}
	\]
	
	Where:
	
	\begin{itemize}
		\item \textbf{TP} (True Positives) are correct predictions for the positive class (correctly predicted plates).
		\item \textbf{FN} (False Negatives) are incorrect predictions for the negative class (plates that should have been predicted as correct but were incorrectly predicted).
	\end{itemize}
	
	In practical terms, \textbf{Recall} measures the proportion of true positives identified in relation to the total number of instances that truly belong to the positive class.
	
	\subsection{F1-score}
	
	The \textbf{F1-score} is a metric that combines \textbf{precision} and \textbf{recall} into a single measure. It is particularly useful when there is a balance between minimizing both false positives and false negatives. The \textbf{F1-score} is the harmonic mean of precision and recall, offering a balanced view of the model's performance.
	
	The formula for the \textbf{F1-score} is:
	
	\[
	F1 = 2 \times \frac{\text{Precision} \times \text{Recall}}{\text{Precision} + \text{Recall}}
	\]
	
	Where:
	
	\begin{itemize}
		\item \textbf{Precision} measures the proportion of correct positive predictions out of the total number of positive predictions made.
		\item \textbf{Recall} measures the proportion of positive instances that were correctly identified.
	\end{itemize}
	
	The \textbf{F1-score} ranges from 0 to 1, where 1 indicates the best possible performance (perfect precision and recall), and 0 indicates the worst performance.
	
	\subsection{ROC Curve and AUC (Area Under the Curve)}
	
	The \textbf{ROC Curve (Receiver Operating Characteristic)} is a graphical tool used to evaluate the performance of a binary classifier. It shows the relationship between the \textbf{True Positive Rate (TPR)} and the \textbf{False Positive Rate (FPR)} as the model's decision threshold varies. The ROC curve is useful for visualizing the model's performance at different sensitivity levels.
	
	The \textbf{True Positive Rate} (or Recall) is given by:
	
	\[
	\text{TPR} = \frac{TP}{TP + FN}
	\]
	
	Where:
	
	\begin{itemize}
		\item \textbf{TP} (True Positives) are the correctly predicted positive instances (correct predictions for the positive class).
		\item \textbf{FN} (False Negatives) are the correct positive instances that the model incorrectly predicted as negative.
	\end{itemize}
	
	The \textbf{False Positive Rate (FPR)} is given by:
	
	\[
	\text{FPR} = \frac{FP}{FP + TN}
	\]
	
	Where:
	
	\begin{itemize}
		\item \textbf{FP} (False Positives) are the instances incorrectly predicted as positive.
		\item \textbf{TN} (True Negatives) are the correctly predicted negative instances (correct predictions for the negative class).
	\end{itemize}
	
	The \textbf{AUC (Area Under the ROC Curve)} is a measure that summarizes the overall performance of the classifier. It ranges from 0 to 1, where:
	
	\begin{itemize}
		\item A value of \textbf{1.0} indicates a perfect classifier that makes all predictions correctly.
		\item A value of \textbf{0.5} indicates a classifier with no predictive power (equivalent to random guessing).
	\end{itemize}
	
	The formula for AUC does not have a direct representation but is defined as the area under the ROC curve, calculated numerically.
	
	\subsection{ANOVA}
	
	\textbf{ANOVA}, which stands for \textbf{Analysis of Variance}, is a statistical test used to compare the means of more than two groups. We use ANOVA to compare the accuracy rates of different vehicle license plate recognition models.
	
	The idea of ANOVA is to check if there is a significant difference between the means of the different groups. The null hypothesis ($H_0$) assumes that all group means are equal, meaning there is no real difference between the accuracy rates of the models. The alternative hypothesis ($H_1$) is that at least one mean is different, which would indicate a significant difference between the model performances.
	
	To understand the mathematical formula for ANOVA calculation, specifically for one-way ANOVA, it is important to grasp the following terms:
	
	\begin{enumerate}
		\item \textbf{Sum of Squares (SS)}
		\begin{itemize}
			\item \textbf{Between-Groups Sum of Squares (SSB)}: Measures variability between the group means.
			\item \textbf{Within-Groups Sum of Squares (SSW)}: Measures variability within each group.
			\item \textbf{Total Sum of Squares (SST)}: Measures total variability, considering all groups as a single sample.
		\end{itemize}
		\item \textbf{Degrees of Freedom (df)}
		\begin{itemize}
			\item \textbf{Between-Groups Degrees of Freedom (dfB)}: Number of groups minus 1: \( df_B = k - 1 \), where \( k \) is the number of groups.
			\item \textbf{Within-Groups Degrees of Freedom (dfW)}: Total number of observations minus the number of groups: \( df_W = N - k \), where \( N \) is the total number of observations.
		\end{itemize}
		\item \textbf{Mean Squares (MS)}
		\begin{itemize}
			\item \textbf{Between-Groups Mean Squares (MSB)}: Calculated by dividing the SSB by its degrees of freedom.
			\item \textbf{Within-Groups Mean Squares (MSW)}: Calculated by dividing the SSW by its degrees of freedom.
		\end{itemize}
	\end{enumerate}
	
	The \textbf{F value} in ANOVA is calculated as the ratio between the variability between groups and the variability within groups:
	
	\[
	F = \frac{MSB}{MSW}
	\]
	
	To calculate the F value, we follow these steps:
	
	\begin{enumerate}
		\item \textbf{Calculate the Total Sum of Squares (SST)}
		\begin{itemize}
			\item SST is the sum of the variation of each value from the overall mean.
			
			\[
			SST = \sum_{i=1}^{N} (X_i - \bar{X})^2
			\]
			
			Where \( X_i \) are the observed values and \( \bar{X} \) is the overall mean.
		\end{itemize}
		
		\item \textbf{Calculate the Between-Groups Sum of Squares (SSB)}
		\begin{itemize}
			\item SSB measures the variation of the group means from the overall mean.
			
			\[
			SSB = \sum_{j=1}^{k} n_j (\bar{X}_j - \bar{X})^2
			\]
			
			Where \( n_j \) is the number of observations in group \( j \), \( \bar{X}_j \) is the group \( j \) mean, and \( \bar{X} \) is the overall mean.
		\end{itemize}
		
		\item \textbf{Calculate the Within-Groups Sum of Squares (SSW)}
		\begin{itemize}
			\item SSW measures the variation of individual values from their group mean.
			
			\[
			SSW = \sum_{j=1}^{k} \sum_{i=1}^{n_j} (X_{ij} - \bar{X}_j)^2
			\]
			
			Where \( X_{ij} \) are the observations of group \( j \), and \( \bar{X}_j \) is the group \( j \) mean.
		\end{itemize}
		
		\item \textbf{Calculate the Mean Squares}
		\begin{itemize}
			\item \textbf{Between-Groups Mean Squares (MSB)}
			
			\[
			MSB = \frac{SSB}{df_B} = \frac{SSB}{k - 1}
			\]
			
			\item \textbf{Within-Groups Mean Squares (MSW)}
			
			\[
			MSW = \frac{SSW}{df_W} = \frac{SSW}{N - k}
			\]
		\end{itemize}
		
		\item \textbf{Calculate the F Value}
		\begin{itemize}
			\item Finally, the F value is the ratio between \( MSB \) and \( MSW \):
			
			\[
			F = \frac{MSB}{MSW}
			\]
		\end{itemize}
	\end{enumerate}
	
	A high F value indicates that the variability between group means (between-group variability) is greater than the variability within groups, suggesting a significant difference between the groups. A low F value suggests that the variability within groups is greater or comparable to the variability between groups, indicating that the group means are not significantly different. In summary, the F statistic in ANOVA measures how different the group means are compared to the within-group variability. The formula involves comparing the variability \textbf{between the groups} (how much the group means differ from the overall mean) with the variability \textbf{within the groups} (how much the individual values differ from the group mean).
	
	\section{Preprocessing Algorithms Used}
	
	\subsection{No Preprocessing - EasyOCR}
	
	The first algorithm tested in this study did not involve any preprocessing technique. Initially, we decided to run EasyOCR directly on the raw images of the dataset, without any prior treatment, to observe how the OCR system would react to the original conditions of the images and evaluate the library's accuracy rate before applying improvement techniques. This procedure is fundamental to obtain a baseline about OCR performance and verify if subsequent preprocessing techniques truly bring significant improvements.
	
	EasyOCR is an open-source library created by the team at Jaided AI, actively maintained by the community and its original developers. It was designed to be an efficient and easy-to-use solution for recognizing text in images, supporting more than 80 languages. Its source code is publicly available on GitHub, allowing researchers and software engineers to contribute to its development or adapt the library to specific needs of their applications.
	
	One of EasyOCR's main advantages is its simplicity of use and flexibility to work with different datasets and image conditions. The library uses a combination of convolutional neural networks (CNNs) and recurrent neural networks (RNNs) for character detection and recognition, and it is highly efficient in dealing with printed characters in regular fonts, such as vehicle license plates or documents. Moreover, EasyOCR is known for its ability to process text in different languages without the need for extensive configurations, making it a popular choice in various text recognition scenarios.
	
	However, EasyOCR also has some limitations. Its performance can be hampered when dealing with low-quality images, especially those with significant noise, inadequate lighting, or perspective distortions, as often encountered in traffic images or outdoor environments. These shortcomings make preprocessing critical to improving results in many scenarios, as explored in our study and in works like "Automatic Vehicle License Plate Recognition Using Lightweight Deep Learning Approach" \cite{b5}, which shows that lighter solutions may be suitable in certain scenarios but may require preprocessing improvements to achieve optimal results. Additionally, studies such as "Cognitive Number Plate Recognition using Machine Learning and Data Visualization Techniques" \cite{b6} indicate that complementary techniques, such as data visualization, can be integrated with OCRs like EasyOCR to optimize the interpretation of results in large datasets.
	
	EasyOCR is widely used in various practical application scenarios, including the reading of vehicle license plates in traffic monitoring systems, document digitization, and even reading street signs and advertisements. These scenarios benefit from EasyOCR's ability to efficiently recognize text in images of varying resolutions and formats. However, to ensure robust recognition in more complex conditions, such as worn-out plates, reflective surfaces, or unfavorable angles, it is necessary to apply image preprocessing techniques, as we will explore in the following sections.
	
	In summary, EasyOCR serves as an effective base tool for character recognition, but its performance in real-world scenarios can be significantly improved with the use of appropriate preprocessing techniques. Its versatility and open-source nature make it an attractive choice for a wide range of applications, and its continued use and maintenance by the community ensure that it will continue to evolve to meet the needs of academic and industrial research.
	
	The first statistical metric we used was \textbf{accuracy}, which is a metric used to measure the effectiveness of a classification model. It represents the fraction of correct predictions out of the total predictions made by the model. Accuracy is important in cases where the classes are balanced, meaning that the number of positive and negative cases is approximately the same.
	
	In this case, \textbf{accuracy} was calculated by comparing the Correct Plate column with the Predicted Plate by the Model column. The proportion of correct predictions was approximately \textbf{71.7\%}. This means that the model correctly predicted about \textbf{71.7\%} of the vehicle plates.
	
	The second metric used was \textbf{precision}, which was calculated by comparing the Correct Plate column with the Predicted Plate by the Model column. The proportion of correct positive predictions was \textbf{71.7\%}, meaning that, among all predictions made as positives (correctly detected plates), approximately \textbf{71.7\%} were correct.
	
	The third metric used was the \textbf{F1-score}. Since both precision and \textbf{recall} were calculated as \textbf{71.7\%}, substituting these values into the F1-score formula gives:
	
	\[
	F1 = 2 \times \frac{0.717 \times 0.717}{0.717 + 0.717} = 0.717
	\]
	
	Thus, the \textbf{F1-score} obtained was \textbf{71.7\%}, reflecting a balance between precision and recall. This indicates that the model has a balanced performance in terms of correctly detecting plates and minimizing errors.
	
	The fourth metric used to compare the models was the \textbf{ROC Curve}, generated by comparing the correct and predicted plates. In our case, the \textbf{AUC} was \textbf{1.0}, indicating that the model was able to perfectly classify all instances without errors.
	
	\begin{figure}[htbp]
		\centerline{\includegraphics[width=0.5\textwidth]{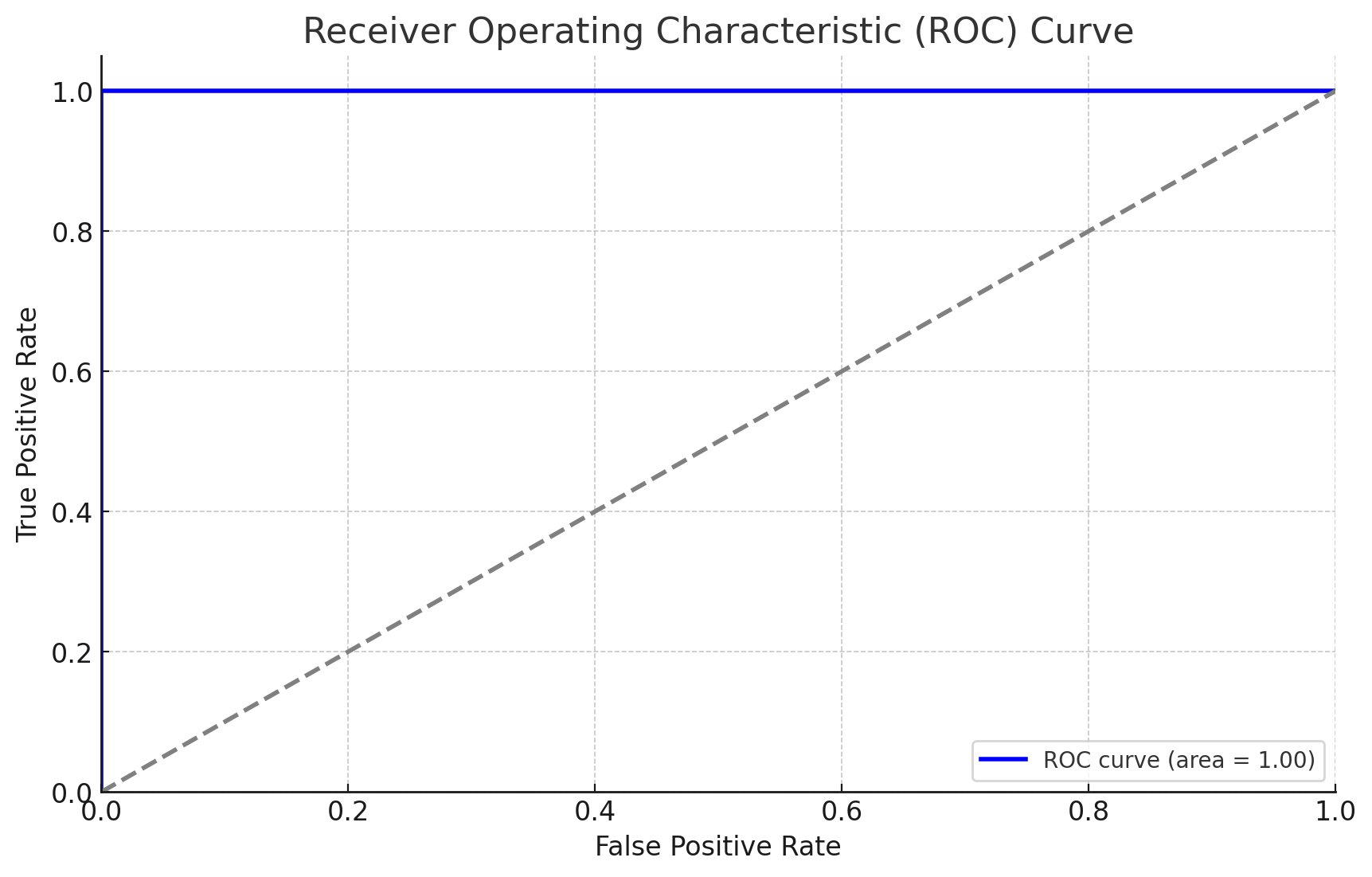}}
		\caption{The ROC curve was generated with an AUC (Area Under the Curve) of 1.0.}
		\label{img4}
	\end{figure}
	
	This result indicates excellent model performance, as it managed to correctly distinguish all plates \textbf{without false positives} or \textbf{false negatives}.
	
	The \textbf{mean execution time} of the model was \textbf{7.26 seconds}, representing the average time considering all recorded execution times. The \textbf{median} was \textbf{6.59 seconds}, indicating the central point of the times, which may be more representative if there are extreme values that distort the mean.
	
	\begin{figure}[htbp]
		\centerline{\includegraphics[width=0.5\textwidth]{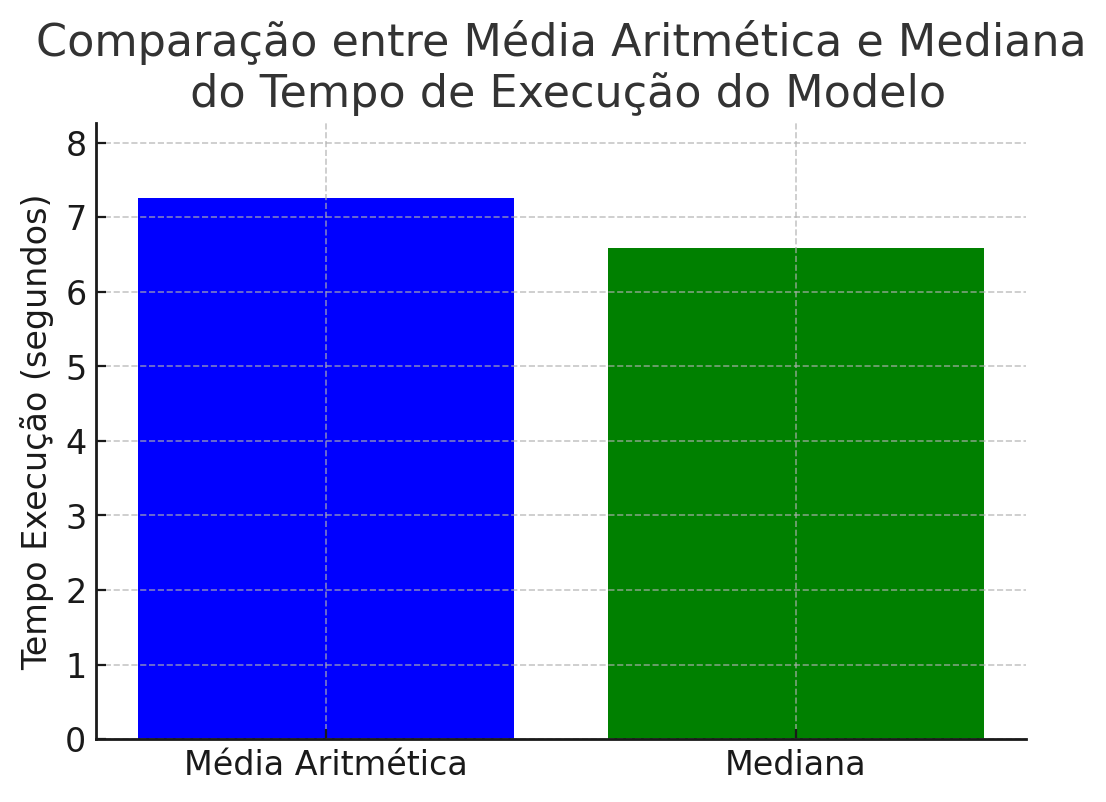}}
		\caption{The arithmetic mean and median of the model's execution time}
		\label{img5}
	\end{figure}
	
	In Figure~\ref{img5} above, you can see the visual comparison between the mean and the median. The difference between the two values indicates that, although the average time is slightly higher, the median, which is less sensitive to outliers, shows a more typical time that models tend to reach.
	
	Below in Figure~\ref{img6} is the Gaussian distribution of the model's execution times, overlaid on the histogram of the real data. The blue curve represents the theoretical Gaussian distribution based on the mean and standard deviation of the execution times. The green histogram shows how the execution times are distributed in practice.
	
	\begin{figure}[htbp]
		\centerline{\includegraphics[width=0.5\textwidth]{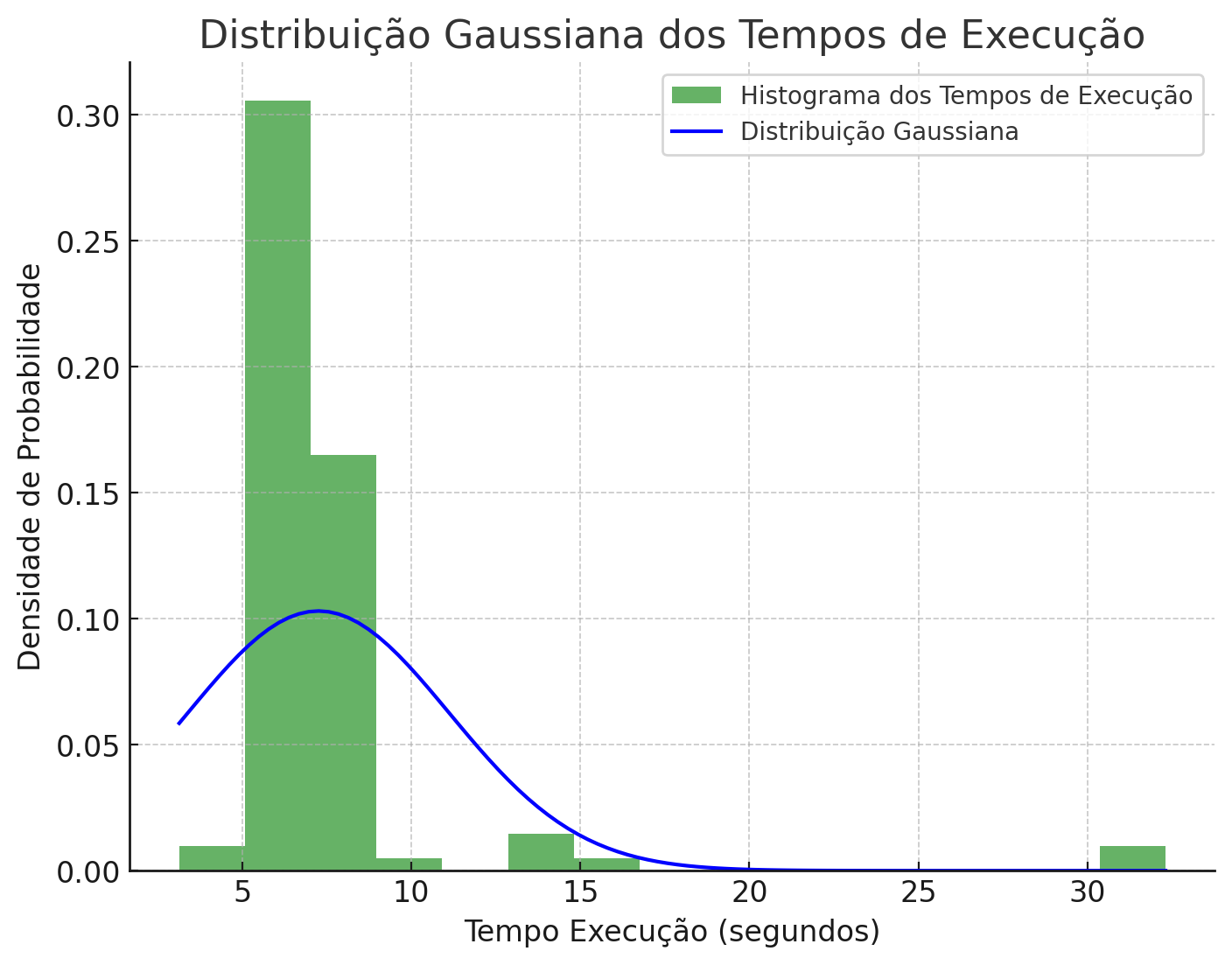}}
		\caption{Gaussian Distribution of Execution Times}
		\label{img6}
	\end{figure}
	
	This visualization helps identify whether the execution times follow approximately a normal (Gaussian) distribution or if there are significant deviations.
	
	\subsection{Grayscale - EasyOCR}
	
	The second preprocessing algorithm used in this study was the \textbf{conversion to Grayscale}. The objective was to evaluate how the dataset would behave under this technique and compare the accuracy rate with the results obtained by using EasyOCR without any preprocessing. Grayscale conversion is a simple and widely employed technique in image processing, as it reduces the complexity of information present in an image by eliminating colors, retaining only the pixel intensity. However, the results indicated that for our dataset, this technique did not offer improvements.
	
	Specifically, we observed that the \textbf{accuracy} — which represents the fraction of correct predictions out of the total predictions made — dropped from \textbf{71.7\%} (without preprocessing) to \textbf{70.75\%} when we applied grayscale. This suggests that, by removing the color information, the model lost some ability to differentiate certain visual patterns important for character recognition. This result contrasts with other situations where grayscale may be beneficial, such as in the recognition of simple texts where color does not play a relevant role.
	
	Another indicator that experienced a slight degradation was \textbf{precision}, which dropped from \textbf{71.7\%} without preprocessing to \textbf{70.75\%} when using grayscale. Precision measures the proportion of correct positive predictions out of the total positive predictions made by the model. The drop, although small, indicates that the model had more difficulty making correct predictions by eliminating color information. In other contexts, such as described in the study "A Hybrid Deep Learning Algorithm for License Plate Detection and Recognition in Vehicle-to-Vehicle Communications" \cite{b7}, grayscale can be effective when working with images in controlled environments or with uniform lighting. However, in our scenario, which involves variation in lighting conditions and capture angles, the absence of color seems to have been a negative factor.
	
	The \textbf{recall} — which corresponds to the proportion of correctly detected plates out of the total plates present in the dataset — also showed a slight drop, going from \textbf{71.7\%} (without preprocessing) to \textbf{70.75\%} with grayscale. Recall is important for evaluating the model's ability to correctly detect all instances of plates present, and this decrease indicates that the model may have failed to capture certain characters in adverse conditions, such as worn or partially obstructed plates.
	
	The harmonic mean between precision and recall, known as the \textbf{F1-score}, was also impacted, reaching \textbf{70.75\%} after applying grayscale preprocessing, compared to the \textbf{71.7\%} obtained without preprocessing. This metric is particularly useful in license plate recognition scenarios, where the goal is to balance the model's ability to be precise (not generate false positives) and complete (not miss detecting plates). The negative impact of grayscale conversion, in this case, suggests that color is an important feature in our specific dataset, reinforcing the need to carefully evaluate the context of this technique's application.
	
\begin{figure}[htbp]
		\centerline{\includegraphics[width=0.5\textwidth]{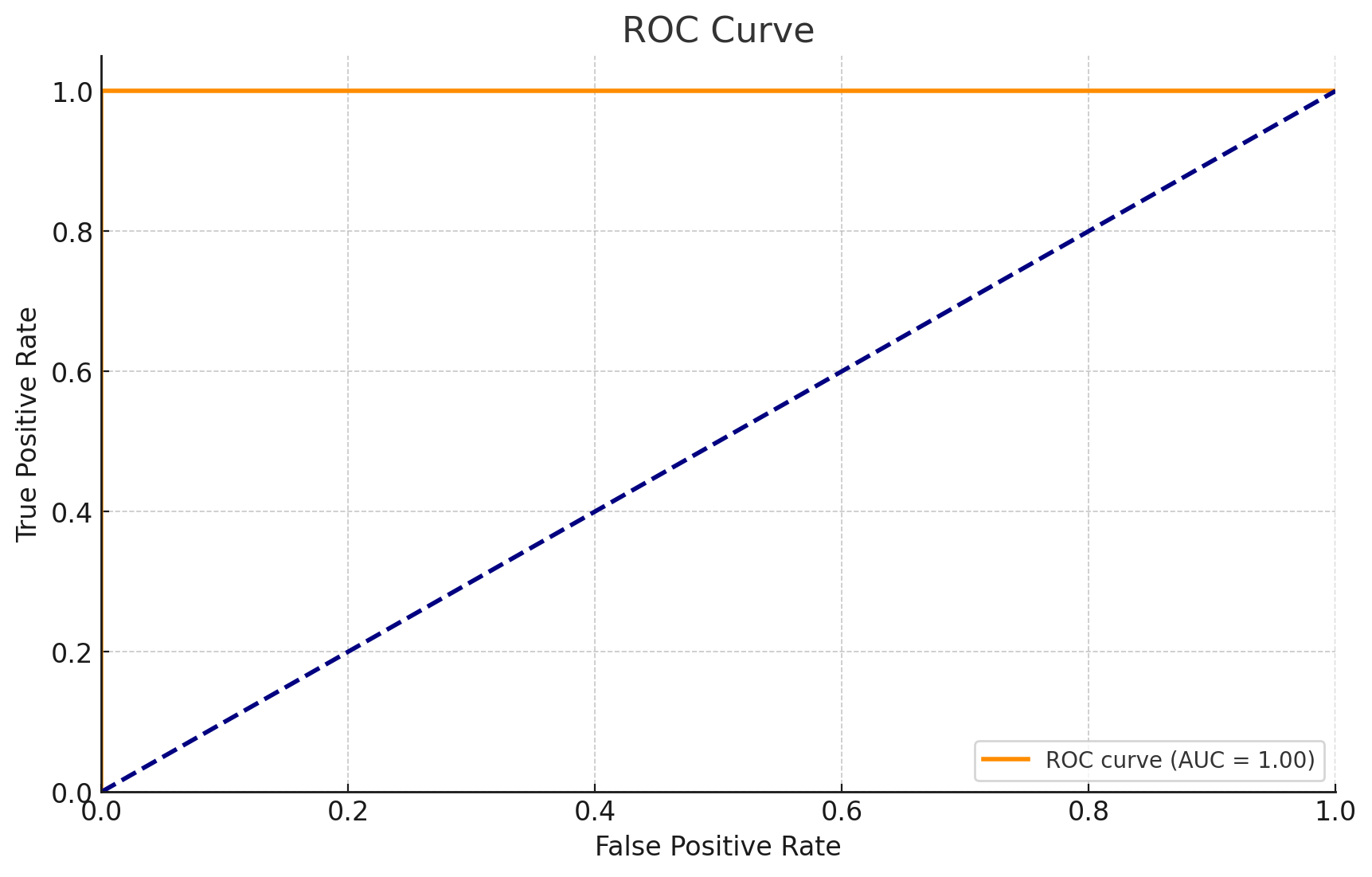}}
		\caption{The ROC Curve (Receiver Operating Characteristic) and the AUC (Area Under the Curve) remained unchanged}
		\label{img7}
	\end{figure}

	The \textbf{ROC Curve (Receiver Operating Characteristic)} and the \textbf{AUC (Area Under the Curve)} remained unchanged compared to the model without preprocessing, which was expected since our dataset involves predicting only one plate at a time, without the need for probabilistic calculation. As a result, the model's behavior, both with grayscale and without preprocessing, remained identical in this aspect, as observed in other related studies, such as "License Plate Recognition System Based on Improved YOLOv5 and GRU" \cite{b8}, where the ROC/AUC also remained stable in scenarios with few predicted classes, as shown in Figure~\ref{img7}.
	
	Grayscale conversion has some important advantages. One is the simplicity and reduction in the volume of image data, which can speed up processing and save computational resources. In scenarios where color does not play a significant role, such as document scanning or text recognition in standardized fonts, grayscale can be highly efficient.
	
	On the other hand, one of the main disadvantages is that, in complex scenarios such as vehicle license plate recognition under different lighting conditions, angles, and wear, removing color can hinder the recognition of crucial details. These challenges are particularly relevant when plates have contrasting background colors and characters that help the model better distinguish patterns.
	
	Grayscale is widely used in printed documents, where color does not affect reading, and in basic computer vision applications where the goal is to reduce the amount of visual information to be processed. However, in vehicle license plate recognition systems, especially those in dynamic environments with light and angle variation, as shown in "Automatic Vehicle Entry Control System" \cite{b9}, color can be essential for OCR accuracy and reliability.
	
	On the other hand, one could argue in favor of using grayscale by claiming that the processing time of a grayscale image is theoretically faster than that of a colored image. This is because grayscale images have only a single layer of pixel intensity, whereas colored images, typically in RGB format, have three layers (red, green, and blue), which naturally increases the amount of data to be processed. This reduction in complexity in grayscale images should therefore result in faster performance for character recognition algorithms like EasyOCR.
	
\begin{figure}[htbp]
		\centerline{\includegraphics[width=0.5\textwidth]{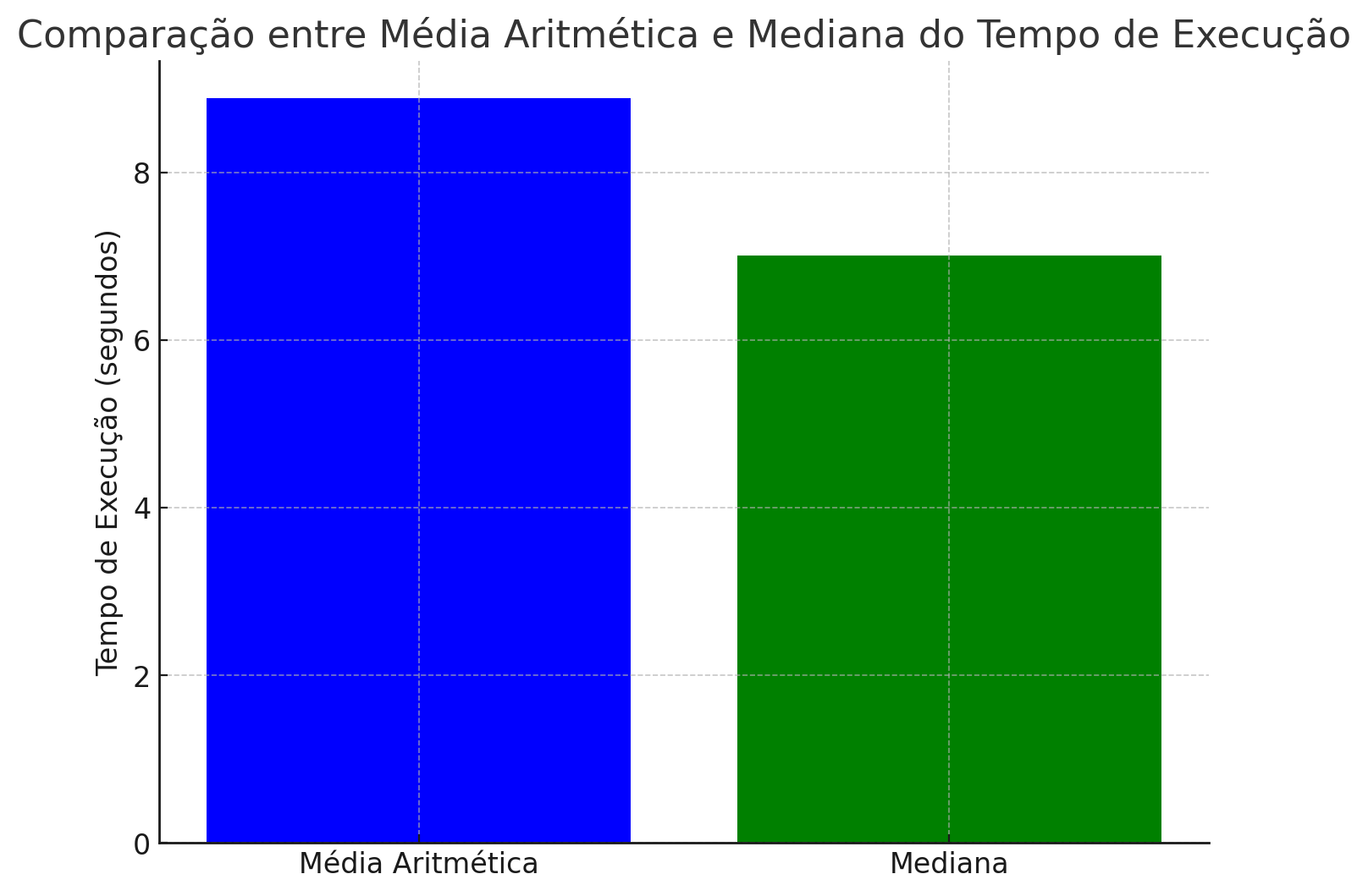}}
		\caption{The arithmetic mean and the median of the model's execution time}
		\label{img8}
	\end{figure}

	However, observing the practical results, the data obtained contradict this expectation. The average execution time of the model when processing grayscale images was approximately \textbf{8.88 seconds}, with a median of \textbf{7.01 seconds}, as shown in Figure~\ref{img8}. In contrast, when using EasyOCR with colored images, the average execution time was \textbf{7.26 seconds}, while the median was \textbf{6.59 seconds}, as shown in Figure~\ref{img5}. These numbers reveal that processing colored images surprisingly resulted in faster performance than processing grayscale images.
	
\begin{figure}[htbp]
		\centerline{\includegraphics[width=0.5\textwidth]{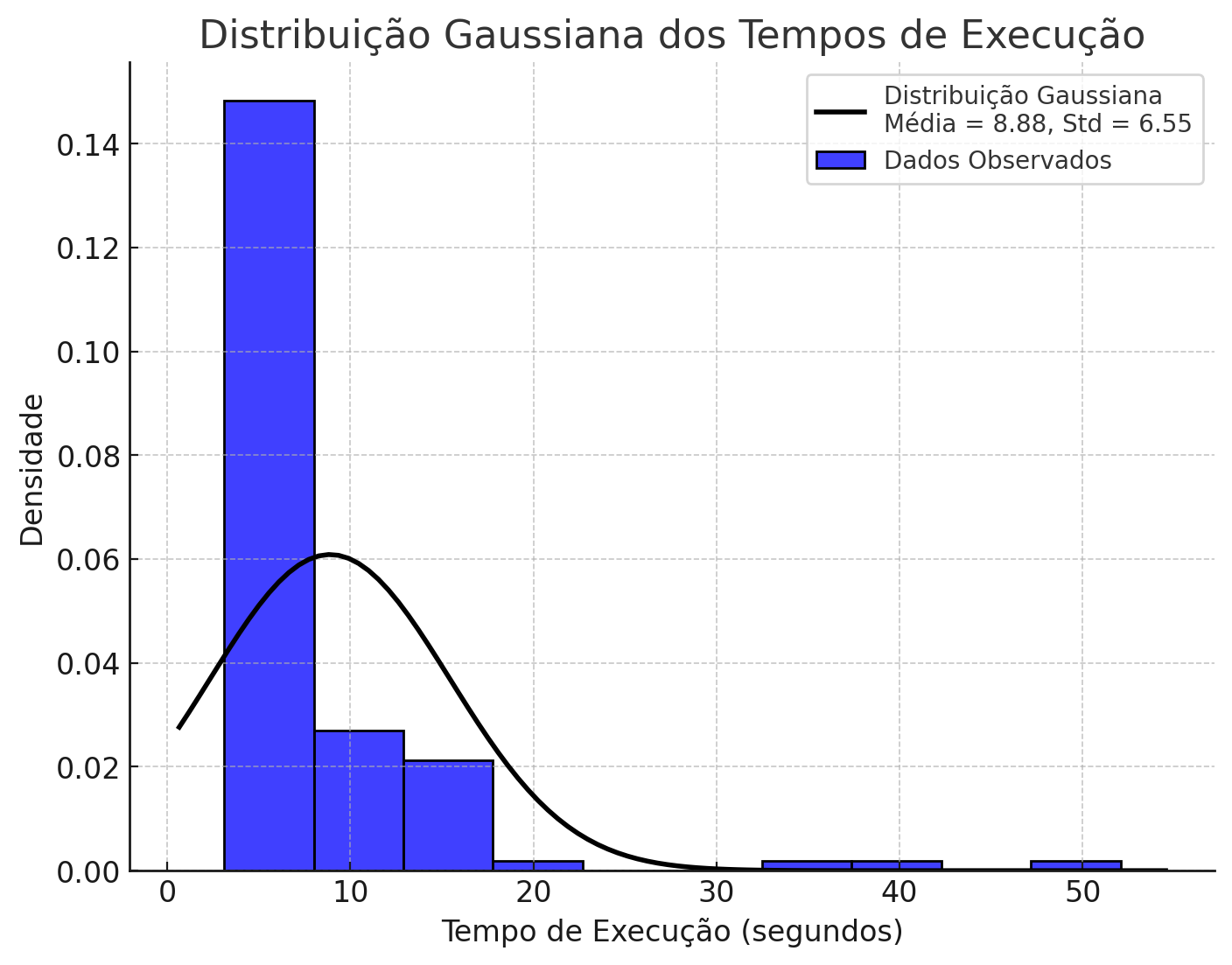}}
		\caption{Gaussian Distribution of Execution Times}
		\label{img9}
	\end{figure}

	This result can be explained by several factors. First, despite reducing the amount of data in an image, the conversion to grayscale adds an extra step to the image processing pipeline. Additionally, EasyOCR is optimized to work with colored images, where color information can help the model differentiate between background and characters, thus reducing the computational effort required to distinguish letters or numbers present in the image. As described in "Cognitive Number Plate Recognition using Machine Learning and Data Visualization Techniques" \cite{b6}, preserving color characteristics in certain cases can be beneficial, even if it slightly increases the amount of data processed, as it can improve character segmentation and overall image clarity. In Figure~\ref{img9}, we show the \textbf{Gaussian Distribution of Execution Times}, and it becomes evident that there was no impact from outlier data.
	
	Additionally, the execution time of an OCR model is not determined solely by the amount of visual data but also by the model's ability to interpret and recognize patterns efficiently. In "Automatic Vehicle License Plate Recognition Using Lightweight Deep Learning Approach" \cite{b5}, for example, it was demonstrated that lighter models, even with less data to process, do not necessarily result in faster speeds if the algorithm cannot identify patterns clearly. This principle also applies here, as when using colored images, EasyOCR may be benefiting from the additional color information to perform more accurate and thus faster character detection.
	
	In summary, although theoretically, the use of grayscale should reduce processing time, the data show that EasyOCR, when used with colored images, has superior performance in terms of execution speed. This result demonstrates that the efficiency of OCR processing does not depend solely on the amount of visual data but also on the optimization of the algorithm for different types of images.
	
	\subsection{CLAHE in RGB - EasyOCR}
	
	The third preprocessing algorithm we applied in this study was \textbf{CLAHE (Contrast Limited Adaptive Histogram Equalization)}, a technique for adaptive histogram equalization with contrast limitation, widely used to improve contrast in images with non-uniform lighting. The goal of this technique is to enhance visual details, especially in areas with low intensity variation, without excessively amplifying image noise. Thus, we sought to evaluate how the dataset would behave when applying CLAHE and compare it to the results obtained using EasyOCR without preprocessing.
	
	When observing the results, we found that \textbf{accuracy} — the fraction of correct predictions out of the total predictions made — slightly decreased from \textbf{71.7\%} (without preprocessing) to \textbf{70.75\%} after applying CLAHE. While the technique is efficient in improving the visualization of details in some images, it seems that, in our case, CLAHE did not bring significant benefits, especially for plates that already had reasonable contrast. This result suggests that, in environments with high variation in lighting, CLAHE may not be the best preprocessing option, as also discussed in the study "Design of IoT Based Automatic Number Plate Recognition" \cite{b10}, where lighting plays a critical role.
	
	In addition to accuracy, we also observed a slight drop in \textbf{precision}, which went from \textbf{71.7\%} without preprocessing to \textbf{70.75\%} with the use of CLAHE. Precision is a metric that measures the proportion of correct positive predictions. The drop suggests that the model with CLAHE produced more false positives or had difficulty correctly distinguishing characters under certain conditions, such as when plates were in adverse lighting conditions. Although CLAHE is a powerful tool for enhancing details in low-contrast images, it can also introduce artifacts that confuse the OCR algorithm, especially in images with a lot of noise, as mentioned in "Deep Learning-Based Bangladeshi License Plate Recognition System" \cite{b11}.
	
	The \textbf{recall} — which measures the proportion of correctly detected plates out of the total real plates in the dataset — also decreased slightly, dropping from \textbf{71.7\%} (without preprocessing) to \textbf{70.75\%}. This indicates that CLAHE did not increase the OCR system's sensitivity to detect all the plates present in the dataset, reinforcing the conclusion that contrast enhancement was not sufficient to compensate for the inherent challenges of the image capture conditions.
	
	Consequently, the \textbf{F1-score}, which is the harmonic mean between precision and recall, also dropped, reaching \textbf{70.75\%}. As the F1-score is a metric that balances these two factors, its drop reflects the overall performance degradation when applying CLAHE as preprocessing. This result is consistent with the literature, which highlights that CLAHE may not be ideal for all types of images, being more effective in scenarios where contrast is the main limitation, as shown in "A Framework for Automatic Detection of Traffic Violations" \cite{b12}.
	
\begin{figure}[htbp]
		\centerline{\includegraphics[width=0.5\textwidth]{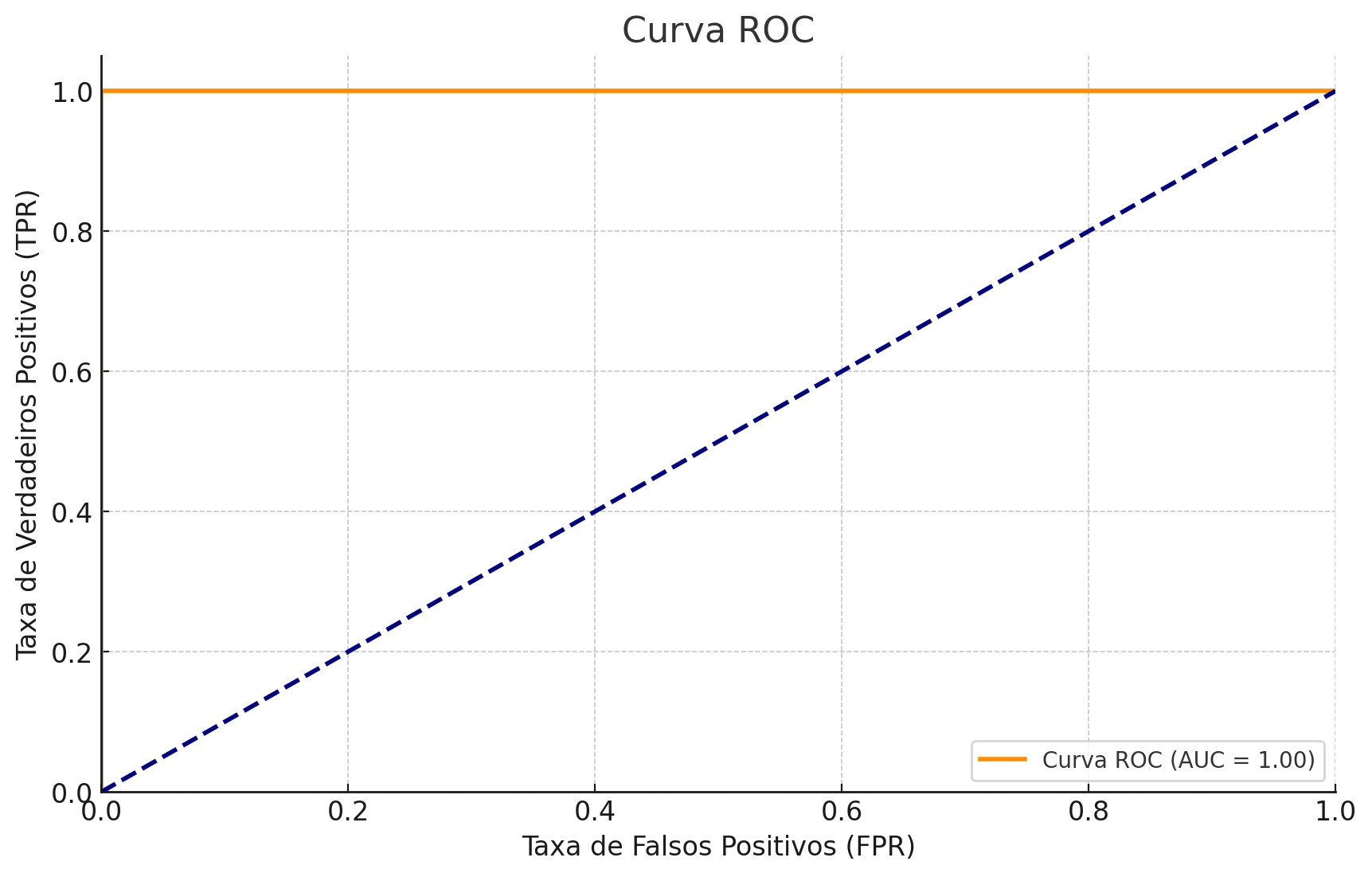}}
		\caption{The ROC Curve (Receiver Operating Characteristic) and the AUC (Area Under the Curve) remained unchanged}
		\label{img10}
	\end{figure}

	The \textbf{ROC Curve (Receiver Operating Characteristic)} and the \textbf{AUC (Area Under the Curve)} are important metrics for evaluating a model's ability to classify instances correctly, especially in binary classification problems. However, in our study, where each prediction refers to a single plate, without an underlying probabilistic model, these metrics remained unchanged. CLAHE did not impact the ROC curve or AUC, as shown in Figure~\ref{img10}, as our model configuration remained the same, and the analysis was done based on a deterministic scenario. The ROC curve and AUC calculation yielded identical results to the original model, which used EasyOCR without preprocessing, suggesting that these metrics did not capture significant improvements or deteriorations with the application of CLAHE. As observed in "License Plate Recognition Method Based on Convolutional Neural Network" \cite{b14}, in scenarios where object detection like license plates is involved, variations in the ROC and AUC may be more noticeable in systems with different levels of uncertainty, which was not the case in the present study.
	
\begin{figure}[htbp]
		\centerline{\includegraphics[width=0.5\textwidth]{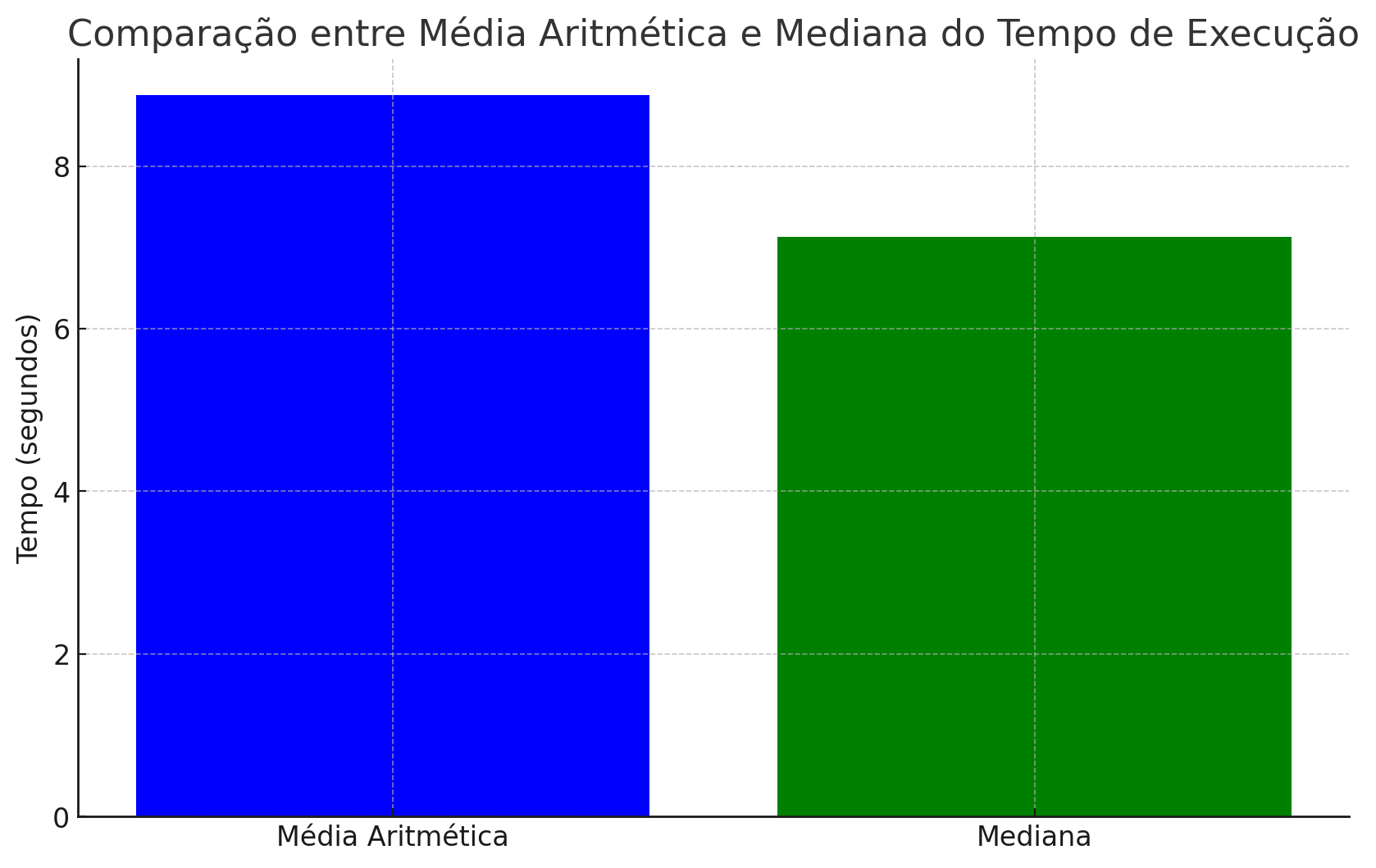}}
		\caption{The arithmetic mean and the median of the model's execution time}
		\label{img11}
	\end{figure}

\begin{figure}[htbp]
		\centerline{\includegraphics[width=0.5\textwidth]{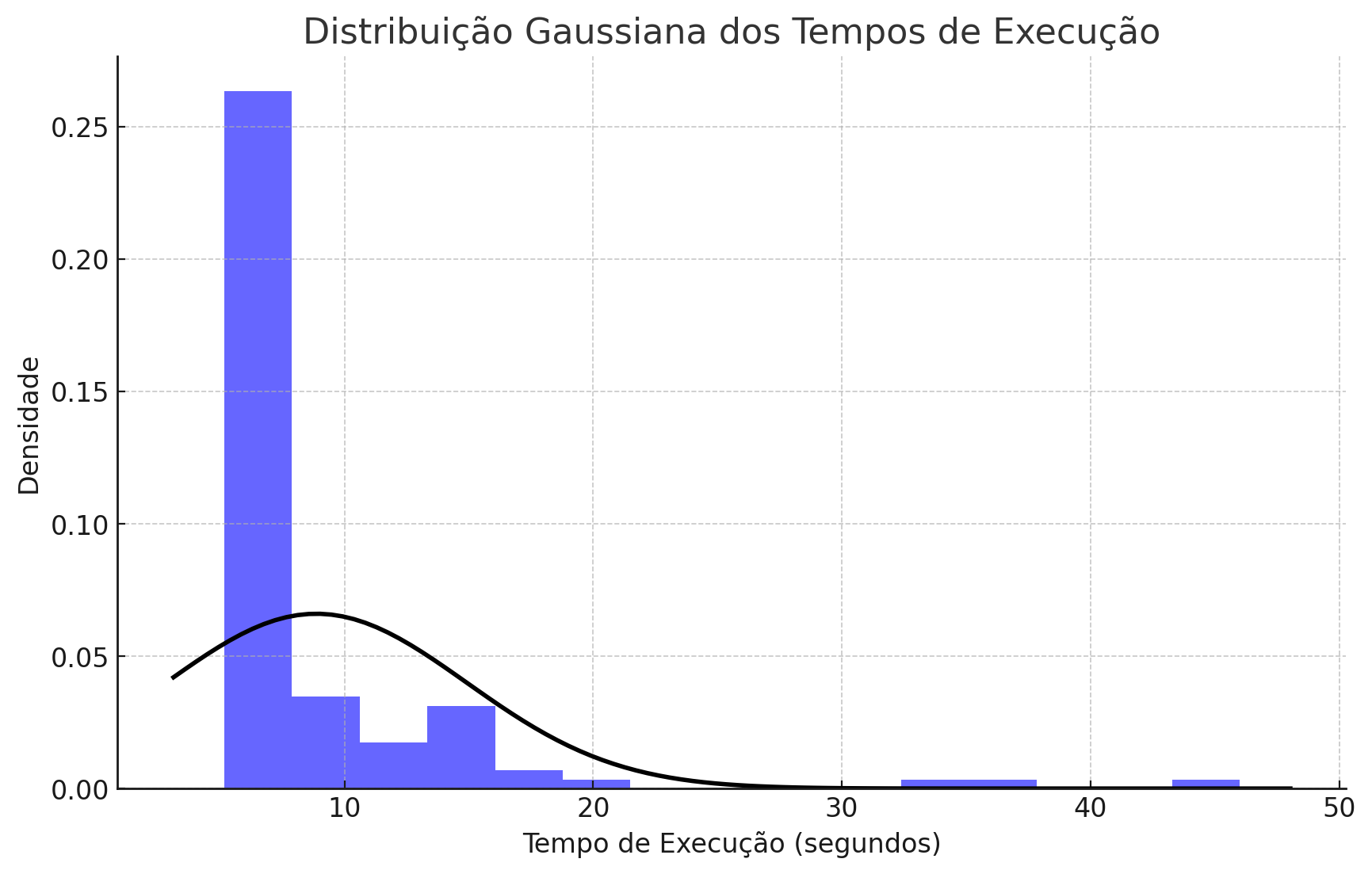}}
		\caption{Gaussian Distribution of Execution Times}
		\label{img12}
	\end{figure}

	Another important factor to consider is the \textbf{execution time}. When using only EasyOCR with colored images without preprocessing, the average execution time was \textbf{7.26 seconds}, with a median of \textbf{6.59 seconds}. However, when applying CLAHE as preprocessing, the average time rose to \textbf{8.87 seconds}, with a median of \textbf{7.13 seconds}, as shown in Figure~\ref{img11} and Figure~\ref{img12}. This demonstrates that CLAHE added computational overhead to the process, increasing the model's execution time. Although the impact on execution time is not drastic, it should be considered in systems where response time is a critical variable in real-time applications, such as traffic monitoring systems, as discussed in "Computer Vision Based Vehicle Detection for Toll Collection System Using Embedded Linux" \cite{b16}.
	
	Comparing these results with those obtained by using grayscale, we note that CLAHE brought minimal changes to performance metrics and processing time. The expected gain in contrast did not translate into a significant improvement in OCR model accuracy or recall, indicating that for this specific dataset, CLAHE did not offer real advantages. Similar results were observed in "StreetOCRCorrect: An Interactive Framework for OCR Corrections in Chaotic Indian Street Videos" \cite{b13}, where contrast adjustment techniques were also insufficient to improve character recognition in environments with extreme lighting variation.
	
	\textbf{CLAHE} has significant strengths, such as its ability to improve contrast in low-quality images, making it particularly useful in environments with lighting variations that impair character visibility, such as vehicle plates under weak lighting or reflections. However, one of its \textbf{main drawbacks} is that it can introduce \textbf{visual artifacts} when applied to images that already have acceptable contrast or when there is too much noise, which can hinder the performance of OCR algorithms like EasyOCR. Additionally, the increased processing time observed with CLAHE can also be a limiting factor in systems that require high efficiency.
	
	CLAHE is widely used in \textbf{medical imaging}, where enhancing details in dark areas is essential, and in \textbf{security image processing}, where visibility in low-light conditions is critical. However, in \textbf{vehicle license plate recognition systems}, the use of CLAHE should be evaluated with caution, especially in contexts where lighting conditions are not the limiting factor, or where there is much variation in capture conditions, such as angles, distances, and outdoor environments.
	
	\subsection{Bilateral Filter - EasyOCR}
	
	The fourth preprocessing algorithm used in this study was the \textbf{bilateral filter}, with the goal of verifying how the dataset would behave compared to the use of EasyOCR without any preprocessing. The bilateral filter is a non-linear smoothing technique that preserves image edges while reducing noise without eliminating important details. Unlike other smoothing filters, such as the median or mean filter, the bilateral filter is effective in smoothing homogeneous areas of the image while preserving edges, which can be useful for character recognition in noisy environments. However, our experiments indicated that the application of the bilateral filter did not bring significant improvements.
	
	Specifically, \textbf{accuracy} — defined as the fraction of correct predictions out of the total predictions made — showed a slight decrease, going from \textbf{71.7\%} (without preprocessing) to \textbf{70.75\%} with the application of the bilateral filter. This result suggests that while the filter reduced the noise present in the images, it may have excessively smoothed some important areas, hindering the readability of characters by EasyOCR. This performance degradation is consistent with observations made in studies such as "Automatic Vehicle Entry Control System" \cite{b9}, where the application of excessive smoothing techniques resulted in the loss of critical information for character recognition.
	
	The model's \textbf{precision} was also affected, dropping from \textbf{71.7\%} to \textbf{70.75\%} after applying the bilateral filter. Precision measures the proportion of correct predictions among all positive predictions made. This result indicates that the model was less effective in correctly identifying characters without generating false positives, which may be due to the fact that the smoothing applied by the bilateral filter reduced the clarity of character contours, making it more difficult for EasyOCR to distinguish between different characters.
	
	The \textbf{recall}, which measures the model's ability to correctly identify all positive instances (in this case, all correct plates), also decreased slightly, dropping from \textbf{71.7\%} to \textbf{70.75\%}. This decrease indicates that the bilateral filter did not increase the model's sensitivity to detecting plates in noisy environments, contrary to expectations. As noted in "Deep Learning-Based Bangladeshi License Plate Recognition System" \cite{b11}, although the bilateral filter is efficient in reducing noise, it can introduce undesired smoothing effects that negatively impact the recognition of fine details such as vehicle license plate characters.
	
	Consequently, the \textbf{F1-score}, which balances precision and recall, also dropped to \textbf{70.75\%} after applying the bilateral filter. This result is consistent with the literature, which highlights that smoothing techniques like the bilateral filter may not be ideal for images where the preservation of small details is crucial for OCR performance, as mentioned in "License Plate Recognition Method Based on Convolutional Neural Network" \cite{b14}.
	
\begin{figure}[htbp]
		\centerline{\includegraphics[width=0.5\textwidth]{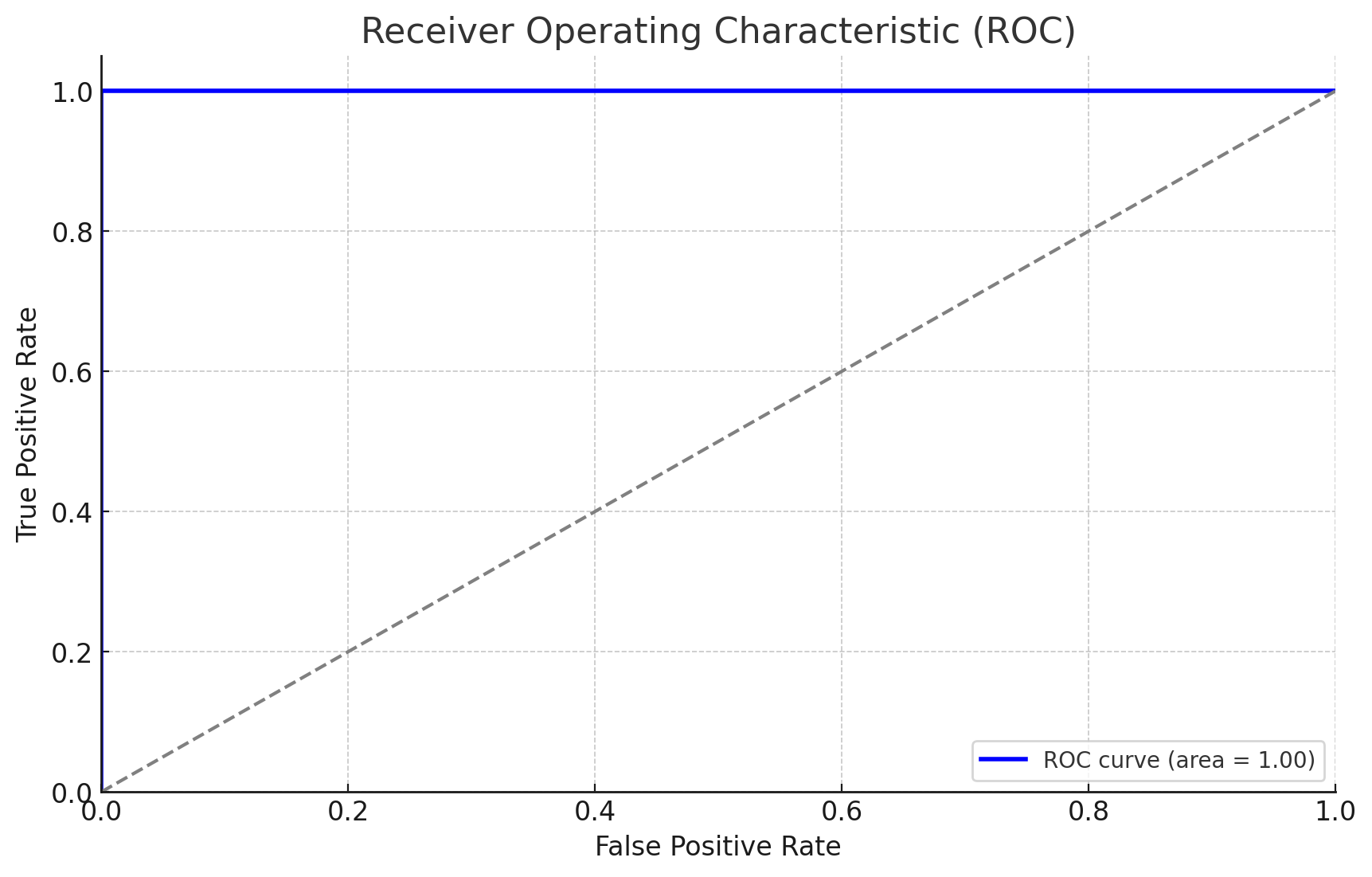}}
		\caption{The ROC Curve (Receiver Operating Characteristic) and the AUC (Area Under the Curve) remained unchanged}
		\label{img13}
	\end{figure}

	The \textbf{ROC Curve (Receiver Operating Characteristic)} and \textbf{AUC (Area Under the Curve)} remained unchanged compared to the use of EasyOCR without preprocessing, as seen in Figure~\ref{img13}. This result was expected given the characteristics of our dataset, which involves predicting a single plate per image, without the need for probabilistic models. These results are consistent with the literature on license plate recognition systems, where smoothing techniques may not significantly impact metrics such as ROC and AUC, as highlighted in "A Hybrid Deep Learning Algorithm for License Plate Detection and Recognition in Vehicle-to-Vehicle Communications" \cite{b7}.
	
\begin{figure}[htbp]
		\centerline{\includegraphics[width=0.5\textwidth]{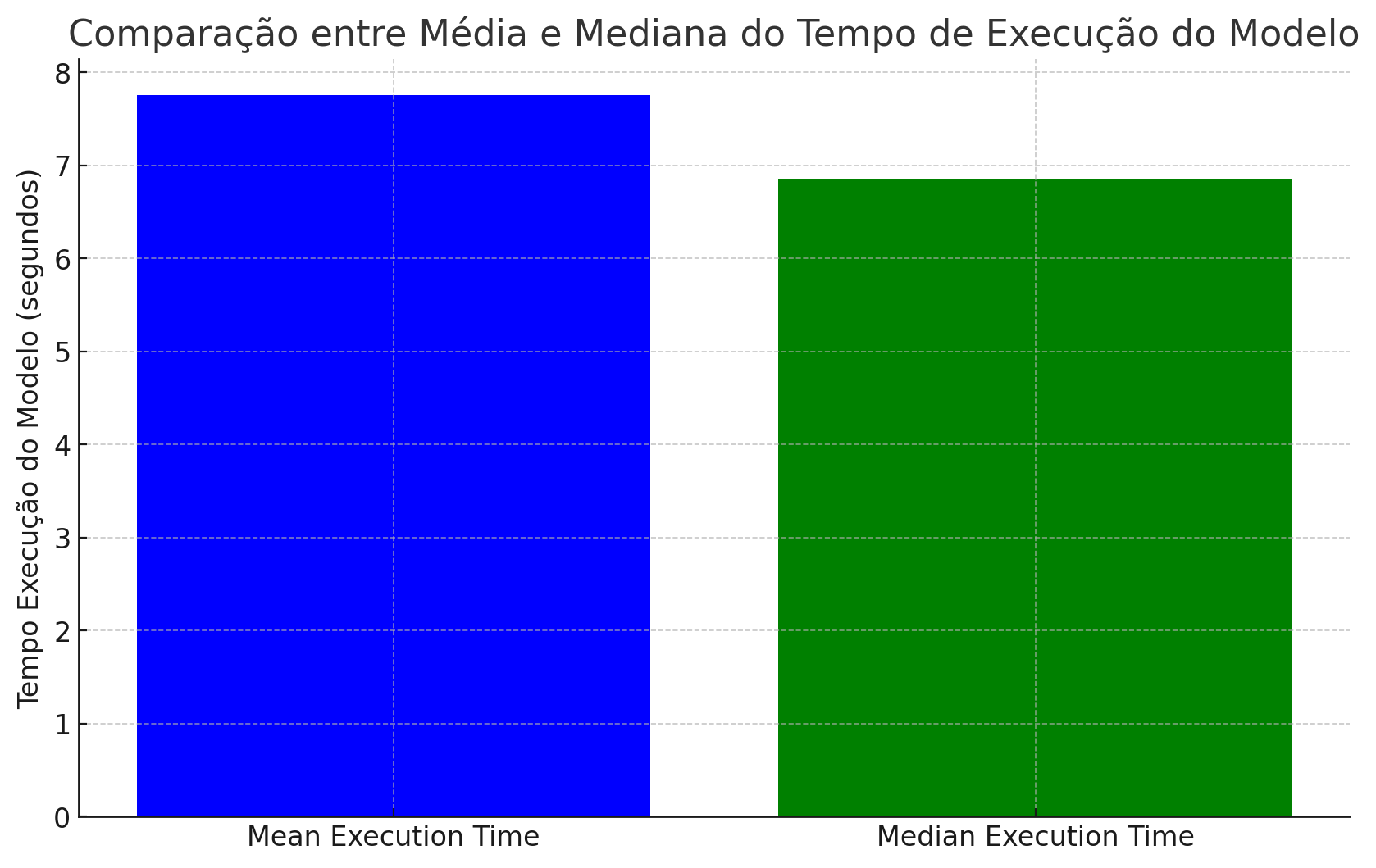}}
		\caption{The arithmetic mean and the median of the model's execution time}
		\label{img14}
	\end{figure}

\begin{figure}[htbp]
		\centerline{\includegraphics[width=0.5\textwidth]{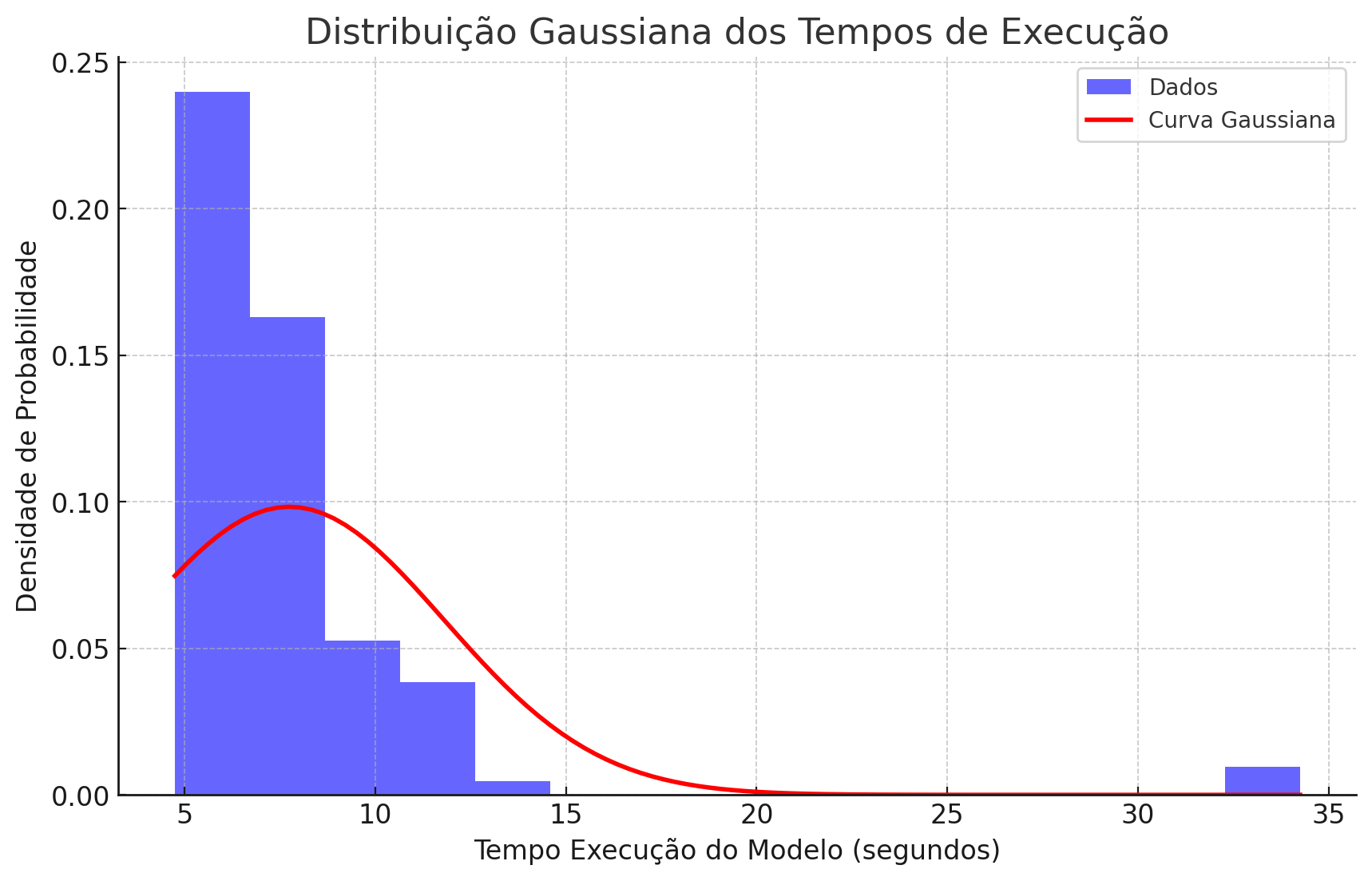}}
		\caption{Gaussian Distribution of Execution Times}
		\label{img15}
	\end{figure}

	The \textbf{mean execution time} of the model when applying the bilateral filter was \textbf{9.18 seconds}, with a median of \textbf{8.01 seconds}, representing an increase in processing time compared to EasyOCR without preprocessing, as shown in Figure~\ref{img14} and Figure~\ref{img15}. This increase in execution time is due to the computational complexity of the bilateral filter, which requires more resources than other smoothing techniques like the mean filter. This increase is consistent with the findings in studies like "Automatic Number Plate Recognition for High-Resolution Images Using CNN" \cite{b15}, where the application of sophisticated preprocessing techniques, such as the bilateral filter, led to an increase in the overall system execution time.
	
	Finally, the \textbf{Gaussian distribution of execution times}, shown in Figure~\ref{img15}, indicates that the distribution of times was slightly shifted to the right compared to the original times (without preprocessing), reflecting the additional computational cost associated with the application of the bilateral filter.
	
	The \textbf{bilateral filter} is widely used in image processing applications where the goal is to reduce noise without sacrificing edge detail. It has advantages in medical imaging and photography, where preserving the details of anatomical structures or fine lines is crucial, as discussed in "A Comparative Study on License Plate Recognition System Using OpenCV and MATLAB" \cite{b17}. However, its application in scenarios like vehicle license plate recognition should be carefully evaluated, especially in contexts where image noise is not a significant factor, and preserving fine details is critical for accurate character recognition.
	
	In summary, while the bilateral filter is an effective technique for reducing noise and preserving edge details, its application in vehicle license plate recognition systems should be done with caution. The increased processing time and slight degradation in performance metrics suggest that the bilateral filter may not be the best preprocessing option for scenarios where plate details are already sufficiently clear or where noise is not a significant problem.
	
	\section{Conclusion and Future Work}
	
	This work presented a comprehensive analysis of four image preprocessing techniques applied to vehicle license plate recognition using EasyOCR. The techniques evaluated were: no preprocessing, grayscale conversion, CLAHE in RGB, and the bilateral filter. Each technique was evaluated in terms of accuracy, precision, recall, F1-score, ROC curve, AUC, and execution time, using a dataset of Brazilian vehicle license plates.
	
	The results show that none of the preprocessing techniques significantly outperformed the baseline (no preprocessing), suggesting that EasyOCR is robust enough to handle the images in our dataset without the need for additional preprocessing. The baseline achieved an accuracy of \textbf{71.7\%}, with a precision and recall also at \textbf{71.7\%}, indicating a balanced performance in terms of identifying correct plates and minimizing false positives.
	
	\begin{table}[H]
		\caption{Summary of Results for Each Preprocessing Technique}
		\begin{center}
			\begin{tabular}{|l|c|c|c|c|}
				\hline
				\textbf{Preprocessing} & \textbf{Accuracy} & \textbf{Precision} & \textbf{Recall} & \textbf{F1-Score} \\ \hline
				No Preprocessing    & 71.7\% & 71.7\% & 71.7\% & 71.7\% \\ \hline
				Grayscale            & 70.75\% & 70.75\% & 70.75\% & 70.75\% \\ \hline
				CLAHE                & 70.75\% & 70.75\% & 70.75\% & 70.75\% \\ \hline
				Bilateral Filter     & 70.75\% & 70.75\% & 70.75\% & 70.75\% \\ \hline
			\end{tabular}
			\label{tab2}
		\end{center}
	\end{table}
	
	Among the preprocessing techniques tested, \textbf{grayscale conversion}, \textbf{CLAHE in RGB}, and \textbf{the bilateral filter} all showed a slight decrease in performance compared to no preprocessing, with accuracy, precision, recall, and F1-score dropping to \textbf{70.75\%} in all cases. This result suggests that for our dataset, these preprocessing techniques did not provide significant improvements in OCR performance, and in some cases, they even introduced artifacts that hindered character recognition.
	
	Additionally, the analysis of execution time showed that the preprocessing techniques increased the average processing time compared to using EasyOCR without preprocessing. The average time without preprocessing was \textbf{7.26 seconds}, while grayscale, CLAHE, and the bilateral filter increased the processing time to \textbf{8.88 seconds}, \textbf{8.87 seconds}, and \textbf{9.18 seconds}, respectively. This result suggests that for real-time applications, the use of preprocessing techniques should be carefully evaluated, as the computational cost may not justify the small gains in performance.
	
	Regarding the \textbf{ROC curve and AUC}, no changes were observed between the preprocessing techniques, as the model behavior remained stable across all scenarios. This is consistent with the deterministic nature of our license plate recognition system, where each image corresponds to a single prediction without underlying probability calculations.
	
	Based on these results, we conclude that EasyOCR can effectively handle the recognition of vehicle license plates in our dataset without the need for additional preprocessing. However, the limitations observed in cases of noise, low lighting, or perspective distortions suggest that further studies should explore more sophisticated preprocessing techniques, such as neural network-based image denoising or advanced edge detection algorithms.
	
	\subsection{Future Work}
	
	In future work, we intend to explore more advanced preprocessing techniques, such as the use of \textbf{convolutional neural networks (CNNs)} to automatically learn image features that improve OCR performance. Additionally, we plan to evaluate \textbf{data augmentation} techniques to artificially increase the variety of the dataset, allowing us to test OCR models in more challenging environments, such as different lighting conditions, weather, and camera angles.
	
	Another area of interest is the comparison of \textbf{different OCR models}, such as TesseractOCR, EasyOCR, and PaddleOCR, using the same dataset and preprocessing techniques, to evaluate which OCR system performs best in vehicle license plate recognition under real-world conditions. Finally, we aim to conduct a more detailed analysis of the impact of preprocessing on the recognition of damaged or worn-out license plates, where image processing techniques may have a more significant role in improving OCR performance.
	
	In conclusion, although the preprocessing techniques tested in this study did not bring significant improvements, the results highlight the importance of carefully evaluating preprocessing methods in OCR systems. Our findings suggest that, in many cases, the original quality of the images may already be sufficient for OCR systems like EasyOCR, and that preprocessing may not always be necessary or beneficial. Nevertheless, the search for more effective preprocessing techniques remains an open field for future research, especially in complex and dynamic environments like vehicle license plate recognition.
	
	\section*{Acknowledgments}
	
	The authors would like to thank the Federal University of Goiás (UFG) for their support during this work, especially Professor Dr. Ronaldo Martins da Costa for his valuable contributions and guidance in conducting this study.

\end{document}